\documentclass[runningheads]{llncs}
\usepackage{graphicx}
\usepackage{tikz}
\usepackage{comment}
\usepackage{amsmath,amssymb} % define this before the line numbering.
\usepackage{color}
\usepackage{hyperref}
\usepackage{array}
\usepackage{booktabs}
\usepackage{multirow}
\usepackage{subfigure}

\usepackage[accsupp]{axessibility}  % Improves PDF readability for those with disabilities.

\begin{document}
% \linenumbers
\pagestyle{headings}
\mainmatter
\def\ECCVSubNumber{5938}  % Insert your submission number here

\title{Learning Causal Features for Incremental Object Detection} % Replace with your title

\begin{comment}
%\titlerunning{ECCV-22 submission ID \ECCVSubNumber}
%\authorrunning{ECCV-22 submission ID \ECCVSubNumber}
%\author{Anonymous ECCV submission}
%\institute{\ECCVSubNumber}
\end{comment}
%******************

\titlerunning{Abbreviated paper title}
\author{Zhenwei He \and Lei Zhang}
\authorrunning{F. Author et al.}
\institute{}
%******************
\maketitle

\begin{abstract}
Object detection limits its recognizable categories during the training phase, in which it can not cover all objects of interest for users. To satisfy the practical necessity, the incremental learning ability of the detector becomes a critical factor for real-world applications. Unfortunately, neural networks unavoidably meet catastrophic forgetting problem when it is implemented on a new task. To this end, many incremental object detection models preserve the knowledge of previous tasks by replaying samples or distillation from previous models. However, they ignore an important factor that the performance of the model mostly depends on its feature. These models try to \emph{rouse} the memory of the neural network with previous samples but not to prevent forgetting. To this end, in this paper, we propose an \textbf{i}ncremental \textbf{c}ausal \textbf{o}bject \textbf{d}etection (ICOD) model by learning causal features, which can adapt to more tasks. Traditional object detection models, unavoidably depend on the data-bias or data-specific features to get the detection results, which can not adapt to the new task. When the model meets the requirements of incremental learning, the data-bias information is not beneficial to the new task, and the incremental learning may eliminate these features and lead to forgetting. To this end, our ICOD is introduced to learn the causal features, rather than the data-bias features when training the detector. Thus, when the model is implemented to a new task, the causal features of the old task can aid the incremental learning process to alleviate the catastrophic forgetting problem. We conduct our model on several experiments, which shows a causal feature without data-bias can make the model adapt to new tasks better.
\end{abstract}

\section{Introduction}

As a basic computer vision task, object detection identifies and locates the object with semantic information of images or video frames. Due to the variety of scenes, viewpoints, equipment, obstacles, and other factors, object detection is a challenging task which attracts amounts of attention in recent years. With the effort of the researchers, object detection has achieved a great process based on CNN in recent years~\cite{he2018mask,ren2017faster,liu2016ssd,fu2017dssd,lin2017focal}.

However, most object detection models limit their recognizable categories during the training, which leads to the gap between the detector and application scenarios. Due to the restriction of the training phase, the traditional detector can not satisfy all application scenes. In fact, the user often requires the detector to incrementally learn new knowledge and expand its application scenarios. That is, when the detection is implemented with a new task, it must perform well on all tasks learned before. However, something is inadequate. When a detector is trained to adapt to a new task, they often fail to preserve the performance of the previous one. The so-called catastrophic forgetting problem~\cite{goodfellow2013empirical,mccloskey1989catastrophic} restrains the incremental learning capability of the detector. Obviously, a straightforward way to solve the catastrophic forgetting is to simultaneously train all tasks, but this kind of method may not work in practice. First, for some privacy reason, data of the old tasks can not be achieved. The joint training of all tasks is sometimes unprocurable. Second, even we can get the data for old tasks, training the model with data of several tasks is time-consuming. To this end, several incremental object detection models are introduced to overcome the problem of catastrophic forgetting~\cite{peng2020faster,shmelkov2017incremental,li2019rilod}, which aims to alleviate the influence of catastrophic forgetting when the model is adapted to a new task.

\begin{figure}[t]
\centering
\includegraphics[width=1.0\linewidth]{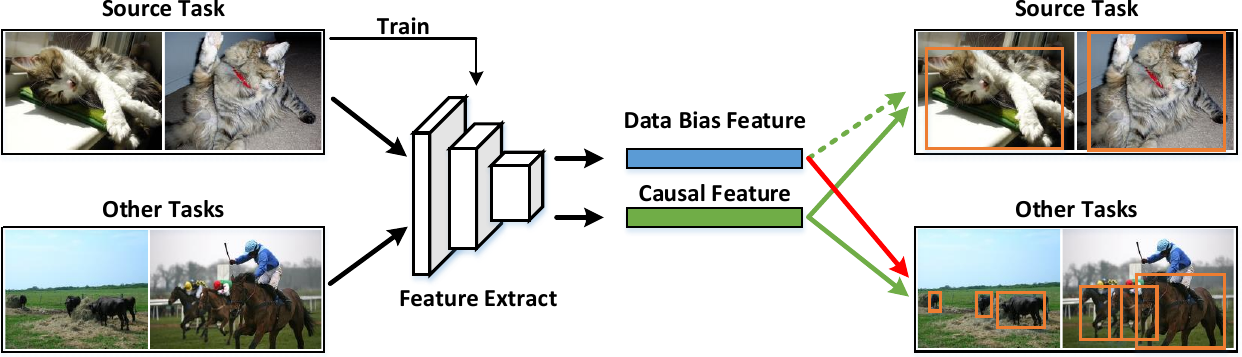}
\caption{The motivation of the paper. For the original task (Source task), when training, both the data bias feature and the causal feature can minimize the loss function. But only the causal features have a reliable relationship to the detection results. For other tasks, only the causal features are useful. By reducing the influence of bias features and enhancing the causal feature, our model can achieve better adaptation to other tasks, such that the problem of catastrophic forgetting is alleviated.}
\label{intro}
\end{figure}

Most of the incremental object detection models use knowledge distillation to alleviate the problem of catastrophic forgetting. In other words, these models simultaneously learn from the previous model and new data in the training phase. But reviewing with previous models brings an amount of computation cost during the training phase. Due to the lack of previous data, the model can not get proper previous knowledge during the training phase. Moreover, we think these methods ignore the analysis of the features, as the feature is the most important factor for the detection. In fact, because of the data bias~\cite{tommasi2017deeper}, the traditional detector learns amounts of data-bias features to improve the performance, which can not adapt to the new task. Although the data-bias feature can reduce the loss functions in the training phase, it does not have a reliable relationship (a.k.a spurious correlation) to the detection results. If a detector depends too much on this kind of bias feature, it is hard to adapt to the new task, such that the catastrophic forgetting problem becomes more serious. In other words, if the features learned by a detector are not favored to the trained data, they can be used for other tasks, and the problem of catastrophic forgetting might be alleviated.

Thus, a detector that can adapt to different tasks should not depend too much on the data bias feature, but on some kind of general features without data bias. Inspired by the causal representation learning~\cite{zhang2021learning,wang2021causal}, which aims to provide the most relative and confident feature representation and reduce the influence of data bias information to the model. In this paper, we introduce a \textbf{i}ncremental \textbf{c}ausal \textbf{o}bject \textbf{d}etection (ICOD) model by learning and extracting causal feature while suppressing the bias feature. When training a detector, as shown in Fig.~\ref{intro}, both the causal features and data bias representation can be learned, but only the causal feature implies as the real relationship to the detection results, while the bias features are just beneficial to get a smaller loss function for the source task. For other tasks, the causal representations are useful while the bias feature is useless due to the dilemma of spurious correlation. Therefore, when the model is implemented on several tasks with incremental learning, it must abandon the data bias information and preserve the causal features. To go further, if the detector mostly depends on the causal features to get the detection results, it is easier for the model to adapt to other tasks. Therefore, different from other methods that rely on old models or data, our ICOD focuses on the training process of the source task, and learns and enhances the causal features to achieve adaptation to other tasks.

In order to train the detector to suppress the data bias feature and enhance the effectiveness of the causal feature, we consider the following two points: First, in order to reduce the influence of data bias features, the model should cut down the relationship between the detection results and the data bias feature. In other words, if the data bias feature does not benefit or even hurt the training phase, the detector will ignore or suppress these features. Second, we think the causal feature is adequate for the model to get a great performance. The detector should have a similar performance with or without bias features. With the above consideration, we design a framework with adversarial learning to locate the data bias features while enhancing the effectiveness of the causal features. In summary, our incremental causal object detection (ICOD) has the following contributions:
\begin{itemize}
\item Different from traditional incremental object detection models, which review the previous knowledge based on the old model or data, our model alleviates the catastrophic forgetting problem by reducing the influence of data bias features and enhancing the causal features. Because the causal features can be used in different tasks, the detector based on the causal features has more adaptability for incremental learning.
\item To enhance the effectiveness of causal features and suppress the data bias, we propose a new training strategy based on adversarial learning. By cutting down the relation of data bias features and the proper detection results, our detector focuses on the causal features to get the detection results, such that the model can adapt to more tasks. Thus, the problem of catastrophic forgetting is alleviated.
\item We implement our model not only on the traditional incremental object detection task with different categories but also test our model on different domains in experiments. The experiment proves the effectiveness of our model.
\end{itemize}
In the following part, we will introduce our incremental causal object (ICOD) detection model in detail, including the details of the model, and corresponding experiments and analysis.

\section{Related Work}

In this part, we will introduce some related work of our model, including object detection models, incremental learning, and causal representation learning.

\subsection{Object Detection}

Object detection is a basic computer vision tasks, which has been studied for a long time~\cite{he2018mask,ren2017faster,fu2017dssd,kim2020cua}. In early age, the object detection is implemented with slide windows and classifier~\cite{dalal2005histograms,felzenszwalb2009object}. Inspired by the development of CNN, recent object detection models achieve a great process. Firstly, the two-stage object detectors are introduced with region of interests (ROIs)~\cite{he2018mask,ren2017faster,girshick2015fast,cai2018cascade}. R-CNN~\cite{girshick2014rich} is one of the earliest models based on CNN, which find the object by classifying the ROIs. After that, Faster-RCNN introduces the RPN for getting the ROIs. Cascade R-CNN~\cite{cai2018cascade}, uses multiple RPN to get precise ROIs. One-stage detectors achieve their success with high detection speed and accuracy~\cite{fu2017dssd,liu2016ssd,redmon2016you}. SSD~\cite{liu2016ssd} finds the objects of different scales with different layers. YOLO~\cite{redmon2016you} gets the detection results with high speed. In recent years, anchor-free methods are also introduced for object detection~\cite{law2018cornernet,tian2019fcos}. FCOS~\cite{tian2019fcos} introduces a coordinate regression method for anchor free detection. DETR~\cite{carion2020end} uses transformer to get the detection results. In this paper, we implement Faster-RCNN as the basic detection model.

\subsection{Incremental Learning}

Artificial neural networks suffer the catastrophic forgetting problem on old tasks when a new task is learned~\cite{delange2021continual}. The major challenge for incremental learning is to learn without catastrophic forgetting. Methods for incremental learning can be divided into three types. First, the replay methods focus on replaying some samples of old tasks~\cite{shin2017continual,rebuffi2017icarl,de2021continual}. iCaRL~\cite{rebuffi2017icarl} stores a subset for each category of the old task. CoPE~\cite{de2021continual} combines nearest mean classifier to overcome the forgetting problem. Second, regularization-based methods~\cite{kirkpatrick2017overcoming,lopez2017gradient} freeze the previous knowledge with network parameters or gradients. EWC~\cite{kirkpatrick2017overcoming} implement a weight for each parameter of the model. Zenke \emph{et al.}~\cite{lopez2017gradient} estimate importance weights during the training phase. Last, some models isolate the parameters of the old tasks~\cite{fernando2017pathnet,rusu2016progressive}. PNN~\cite{rusu2016progressive} uses a new branch for new tasks while freezing the old one. PathNet assigns different tasks with the different paths of the network.

\subsection{Causal representation Learning}

Causal representation learning~\cite{glymour2016causal,scholkopf2019causality} has been used for computer vision in recent years. Shalit \emph{et al.}~\cite{shalit2017estimating} proposes a new theoretical analysis and some algorithms for predicting individual treatment effects from observational data. For feature representation, VC RCNN~\cite{Wang_2020_CVPR} correct the errors of visual relationship and get more
accurate visual relationships and visual attention. IFSL~\cite{yue2020interventional} is introduced for few-shot learning based on causal intervention. CONTA~\cite{zhang2020causal} propose a structural causal model to analyze the
causalities among images. In this paper, we propose to enhance the causal features for the detection model, which is profitable for other tasks.

\section{Causal Preliminaries}

\begin{figure}[h]
\centering
\includegraphics[width=0.6\linewidth]{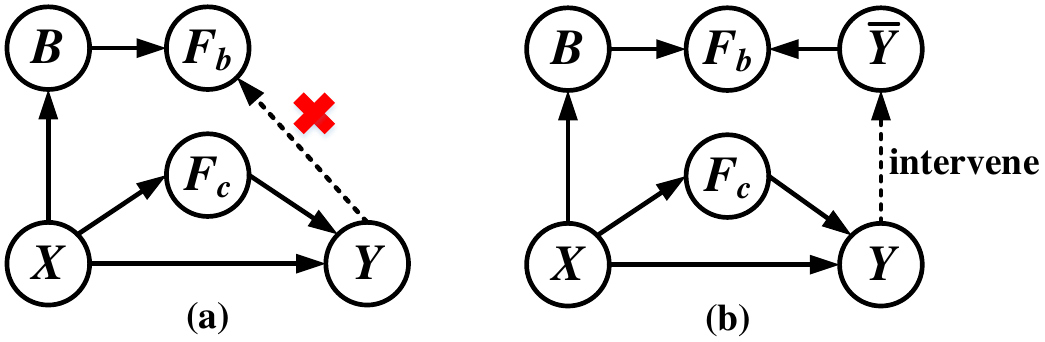}
\caption{The causality graph of the detection model. (a). The traditional object detection model may establish a path from $Y$ to $F_{b}$, which leads to a backdoor path and create a spurious correlation between data bias information $B$ and ground truth $Y$. (b). In our model, we intervene $Y$ and get $\overline{Y}$ to guide the training of $F_{b}$, which sever the path from $Y$ to $F_{b}$. The model only depends on $F_{c}$ to predict the results.}
\label{causal}
\end{figure}

\textbf{The traditional detector.} The causality graph of traditional and our ICOD are shown in Fig.~\ref{causal} (a) and (b), respectively. Some important elements are presented in the figure, including input image $X$, data bias information $B$, data bias features $F_{b}$, causal features $F_{c}$, and the ground truth label $Y$. The direct links show the causalities between the two nodes. In Fig.~\ref{causal}.(a), we show the causal relations of the input image, features, and ground truth label in the traditional object detection model.

\textbf{$X\rightarrow F_{c}\rightarrow Y$} The causal features contain the intrinsic information for the object detection task, which are extracted based on the input image. With the causal features, the model can also deduce the proper detection results after the training phase. We think this kind of feature can be used in different tasks.

\textbf{$X\rightarrow B\rightarrow F_{b}\leftarrow Y$} The dataset unavoidably contains data bias information, which does not have a reliable relationship to the ground truth $Y$. However, data bias features $F_{b}$ can help to minimize the detection loss function, thus, the detection model may establish a path from $Y$ to $F_{b}$ during the training phase. Note that $Y\rightarrow F_{b}$ is an illusory route because the model mistaken these features are important based on the loss functions and backpropagation. Moreover, the illusory route can establish a backdoor path and create a spurious correlation between data bias information $B$ and ground truth $Y$, which makes the decision of the detector depending on the data bias information.

\textbf{Detector of ICOD.} From Fig.~\ref{causal}.(a), the data bias feature $F_{b}$ is learned based on the data bias information $B$ and the ground truth $Y$. To reduce the influence of the data bias feature, we should sever the path between $F_{b}$ and $Y$ during the training phase. Considering that $F_{b}$ does not have reliable relation to $Y$, $F_{b}$ can establish an illusory route to any kind of label. Therefore, we can use a fake ground truth label $\overline{Y}$ to confuse $F_{b}$.

Fig.~\ref{causal}.{b} shows the causality graph of ICOD, where we propose an intervention of $Y$, and get $\overline{Y}$ to guide the training process of data bias feature $F_{b}$. With $\overline{Y}$ to train the $F_{b}$, the path from $Y$ to $F_{b}$ is estimated. Therefore, there is no backdoor collection between $B$ and $Y$ for ICOD. Besides, the only path for the detector to get the ground truth $Y$ is $X\rightarrow F_{c}\rightarrow Y$, which forces the model to focuses on the causal features $F_{c}$. As result, ICOD is trained mainly based on causal features, and can easily adapt to other tasks.

\section{Methods}

As presented before, our ICOD focuses on the causal features by confusing the data bias features with fake label $\overline{Y}$. In this part, we will introduce our incremental causal object detection (ICOD) in detail.

\begin{figure}[t]
\centering
\includegraphics[width=1.0\linewidth]{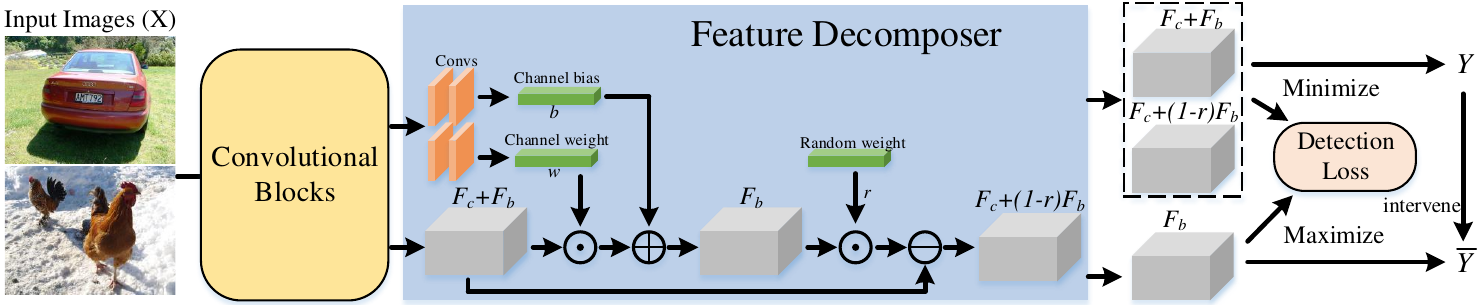}
\caption{The overview of the ICOD. Our model is based on Faster-RCNN. On the top of the convolutional blocks, we implement the feature decomposer to get different kinds of features, including the data bias feature and causal features. After that, the original feature, data bias feature, and causal feature are fed into the detection loss for training.}
\label{overview}
\end{figure}

\subsection{Overview of ICOD}

The overview of ICOD can be observed in Fig~\ref{overview}, which proposes Faster-RCNN~\cite{ren2017faster} as the basic detector. On the top of convolutional blocks, to get the data bias and causal features, we introduce the feature decomposer to decompose the original features. After that, the original features, causal features, and data bias feature are fed into the detection loss (i.e. classification loss and regression loss) for training. The original feature and causal feature aim to minimize the detection loss, while the data bias feature is trained to maximize the loss. After the training of the source task, for the incremental learning process, we use the finetune method or based on EWC~\cite{kirkpatrick2017overcoming} to learn other tasks, which is described in the experiment part in detail.

\subsection{Feature Decomposer}

Feature Decomposer is designed to get the data bias and causal feature during the training phase. Inspired by~\cite{bau2018gan}, which presented that different channels in feature maps stand for different kinds of information, we think the data bias information and causal information exist in the different channels of the original features. In fact, the original feature maps contain both the data bias information and causal information, feature decomposer aims to separate the above two kinds of information.

In detail, we introduce a channel weight and a channel bias to get the data bias feature based on the original features. The channel weight part is composed of two convolution layers and a sigmoid activation function, such that the channel weight is close to 0 or 1 to filter out some channels. Besides the channel filter, we also implement two convolution layers to learn the channel bias. Suppose the net for channel filter can be presented as $N_{f}$, net for channel bias is $N_{b}$, the original feature is $F$. The data bias feature in our model can be presented as:
\begin{align}
F_{b} = w\odot F + b
\label{eq1}
\end{align}
where $w=N_{f}(F)$ and $b=N_{b}(F)$ are the channel weight and channel bias computed by the network, respectively. With Eq.~\ref{eq1}, we can get the data bias feature $F_{b}$. As the original feature $F$ is the combination of $F_{b}$ and $F_{c}$, we can also get $F_{c}$. In order to improve the variety of the feature, we use a random weight to get the causal feature:
\begin{align}
F_{c} = F - r\odot F_{b}
\label{eq2}
\end{align}
where $r$ is the random weight, which follows the uniform distribution from 0 to 1. The random weight makes $F_{c}$ contain some data bias information. In practice, the pure causal features are hard to be achieved. Thus, the random weight $r$ can help the feature to preserve some data bias information to approach the practical condition. Besides, in order to prevent the trivial solution of channel weight and channel bias. We think the channel filter should be sparse, and restrict it with L1 loss. To reduce the scale of the channel bias, it is restricted with L2 loss. Totally, the loss function for channel weight $w$ and channel bias $b$ is:
\begin{align}
L_{w,b} = \alpha\left \| w\right \|_{1} + \beta\left \| b\right \|_{2}^{2}
\label{eq3}
\end{align}
where $\alpha$ and $\beta$ are the hyper-parameters for the loss function. With the restrict of Eq.~\ref{eq3}, the trivial solution is prevented.

\subsection{The loss function of ICOD}

With feature decomposer, our model has three types of features for training the detector, including the original feature, data bias feature, and causal feature. ICOD is trained with all three kinds of features. With the training phase, we hope our model can focus on the causal feature $F_{c}$ with the following consideration: First, both the original feature $F$ and causal feature $F_{c}$ contain all useful information, they should get the proper detection results after the training. So we minimize the detection loss based on $F$ and $F_{c}$. Second, according to the discussion of Section.3, the information contained in the data bias feature does not have any reliable relation to the detection results. Therefore, with data bias features, we maximize the detection loss to lead them to a random $\overline{Y}$. Last, since $F$ and $F_{c}$ are trained to minimize the loss function, features that can be lead to $\overline{Y}$ should be the data bias features.

With the above consideration, suppose the detection loss function is $L_{d}$, which contains the regression loss (SmoothL1 loss) and classification loss (Cross-entropy loss) of RPN and RCNN parts. Suppose $\theta_{m}$ stands for the parameter of the detector backbone, $\theta_{c}$ is the parameter of network for channel weight or bias. The loss function for our ICOD can be written as:
\begin{align}
L_{ICOD} = L_{c}(\theta_{m})+\gamma L_{b}(\theta_{c})+ L_{w,b}(\theta_{c})
\label{eq4}
\end{align}
Eq.~\ref{eq4} has three parts, the first part is $L_{c}=\min_{\theta_{m}}[L_{d}(F, Y)+L_{d}(F_{c},Y)]$, which minimize the detection loss with $F_{c}$ and $F$. The loss function trains the detector with causal features. The second part is $L_{b}=\max_{\theta_{c}}L_{d}(F_{b},Y)$ optimizes the parameter of channel weight and bias parts. The third is from Eq.~\ref{eq3}. The loss function can locate the data bias feature from the original feature $F$, and reduce the influence of the data bias feature during the training.

\subsection{Explanation of the loss function}

In this part, we will explain the effect of the loss function in Eq.~\ref{eq3}. Except for training the detector, the loss function has the following effects: First, by simultaneously minimizing the detection loss based on $F$ and $F_{c}$, the detector can get proper detection results with both $F$ and $F_{c}$, such that the results of $F$ and $F_{c}$ becomes similar. That is, the detector can detect the objects with or without the data bias feature. The effectiveness of the data bias feature is reduced. Second, as discussed in Section.3, maximizing the loss function with $F_{b}$ can establish a new path from $F_{b}$ to $\overline{Y}$, such that the spurious correlation between $B$ and $Y$ is disappeared. Third, by training the model with the loss function with adversarial learning strategy, $F_{b}$ is hard to contain any causal information. That is because any causal information in $F_{b}$ will reduce the effectiveness of $F_{c}$ due to Eq.~\ref{eq2}, which makes the minimization of detection loss with $F_{c}$ much harder.

Thus, by minimizing and maximizing the detection loss, the model can get the data bias feature, and alleviate its influence during the training phase. Finally, our ICOD is trained to only depend on the causal features for the detection.

\section{Experiment}

In this section, we implement our ICOD on the tasks of incremental object detection with several datasets, including Pascal VOC~\cite{everingham2010pascal}, Cityscapes~\cite{cordts2016cityscapes} and Foggy Cityscapes~\cite{sakaridis2018semantic}. Besides the traditional tasks which implement different categories as different tasks, we suppose that different scenes or domains may also cause the forgetting problem. In the experiment part, we implement our model on the two kinds of scenarios. Faster-RCNN~\cite{ren2017faster} is used as the baseline model of our ICOD. First, we will briefly introduce the corresponding datasets for the experiment.

\begin{table}[t]
\begin{center}
\caption{Per-class accuracy of Pascal VOC (\%) for 10+10}
\label{10-10}
\begin{tabular}{l|p{0.8cm}<{\centering}p{0.8cm}<{\centering}p{0.8cm}<{\centering}p{0.8cm}<{\centering}p{0.8cm}<{\centering}p{0.8cm}<{\centering}p{0.8cm}<{\centering}p{0.8cm}<{\centering}p{0.8cm}<{\centering}p{0.8cm}<{\centering}|c}
\hline
Methods & aero & bike & bird & boat & bottle & bus & car & cat & chair & cow & ~ \\
\hline
\hline
A11 20 & 68.5 & 77.2 & 74.2 & 55.6 & 59.7 & 76.5 & 83.1 & 81.5 & 52.1 & 79.8 & ~ \\
First 10 & 79.3 & 79.7 & 70.2 & 56.4 & 62.4 & 79.5 & 88.6 & 76.6 & 50.1 & 68.9 & ~ \\
New 10 & 7.9 & 0.3 & 5.1 & 3.4 & 0 & 0 & 0.2 & 2.3 & 0.1 & 3.3 & ~ \\
\hline
ILOD~\cite{shmelkov2017incremental} & 69.9 & 70.4 & 69.4 & 54.3 & 48.0 & 68.7 & 78.9 & 68.4 & 45.5 & 58.1 & ~ \\
Faster-ILOD~\cite{peng2020faster} & 72.8 & 75.7 & 71.2 & 60.5 & 61.7 & 70.4 & 83.3 & 76.6 & 53.1 & 72.3 & ~ \\
\hline
ICOD(ours) & 77.2 & 53.5 & 65.9 & 56.5 & 22.3 & 65.2 & 42.9 & 79.2 & 23.5 & 67.7 & ~ \\
\hline
\hline
Methods & table & dog & horse & mbike & prsn & plant & sheep & sofa & train & tv & mAP \\\hline
\hline
All 20 & 55.1 & 80.9 & 80.1 & 76.8 & 80.5 & 47.1 & 73.1 & 61.2 & 76.9 & 70.3 & 70.5 \\
First 10 & 0 & 0 & 0 & 0 & 0 & 0 & 0 & 0 & 0 & 0 & 35.6 \\
New 10 & 65.0 & 69.3 & 81.3 & 76.4 & 83.1 & 47.2 & 67.1 & 68.4 & 76.5 & 69.2 & 36.3 \\
\hline
ILOD~\cite{shmelkov2017incremental} & 59.7 & 72.7 & 73.5 & 73.2 & 66.3 & 29.5 & 63.4 & 61.6 & 69.3 & 62.2 & 63.2 \\
Faster-ILOD~\cite{peng2020faster} & 36.7 & 70.9 & 66.8 & 67.6 & 66.1 & 24.7 & 63.1 & 48.1 & 57.1 & 43.6 & 62.2 \\
\hline
ICOD(ours) & 71.6 & 74.1 & 82.7 & 78.2 & 79.1 & 45.8 & 68.8 & 67.6 & 77.9 & 66.2 & \textbf{63.3} \\
\hline
\end{tabular}
\end{center}
\end{table}

\begin{table}[t]
\begin{center}
\caption{Per-class accuracy of Pascal VOC (\%) for 15+5}
\label{19-1}
\begin{tabular}{l|p{0.8cm}<{\centering}p{0.8cm}<{\centering}p{0.8cm}<{\centering}p{0.8cm}<{\centering}p{0.8cm}<{\centering}p{0.8cm}<{\centering}p{0.8cm}<{\centering}p{0.8cm}<{\centering}p{0.8cm}<{\centering}p{0.8cm}<{\centering}|c}
\hline
Methods & aero & bike & bird & boat & bottle & bus & car & cat & chair & cow & ~ \\
\hline
\hline
First 15 & 74.2 & 79.1 & 71.3 & 60.3 & 60.0 & 80.2 & 88.1 & 80.2 & 48.8 & 74.6 & ~ \\
New 5 & 3.7 & 0.5 & 6.3 & 4.6 & 0.9 & 0 & 8.8 & 3.9 & 0 & 0.4 & ~ \\
\hline
ILOD~\cite{shmelkov2017incremental} & 70.5 & 79.2 & 68.8 & 59.1 & 53.2 & 75.4 & 79.4 & 78.8 & 46.6 & 59.4 & ~ \\
Faster-ILOD~\cite{peng2020faster} & 66.5 & 78.1 & 71.8 & 54.6 & 61.4 & 82.6 & 82.7 & 52.1 & 74.3 & 63.1 & ~ \\
\hline
ICOD(ours) & 78.0 & 75.3 & 73.3 & 59.9 & 42.9 & 84.7 & 76.5 & 78.7 & 23.1 & 82.8 & ~ \\
\hline
\hline
Methods & table & dog & horse & mbike & prsn & plant & sheep & sofa & train & tv & mAP \\\hline
\hline
First 15 & 61.0 & 76.0 & 85.3 & 78.2 & 83.4 & 0 & 0 & 0 & 0 & 0 & 55.0 \\
New 5 & 0 & 0 & 16.4 & 0.7 & 0 & 41.0 & 55.7 & 49.2 & 59.1 & 67.8 & 15.9 \\
\hline
ILOD~\cite{shmelkov2017incremental} & 59.0 & 75.8 & 71.8 & 78.6 & 69.6 & 33.7 & 61.5 & 63.1 & 71.7 & 62.2 & 65.9 \\
Faster-ILOD~\cite{peng2020faster}  & 63.1 & 78.6 & 80.5 & 78.4 & 80.4 & 36.7 & 61.7 & 59.3 & 67.9 & 59.1 & 67.9 \\
\hline
ICOD(ours) & 61.6 & 74.3 & 82.5 & 76.5 & 65.0 & 49.2 & 74.2 & 66.9 & 83.8 & 59.4 & \textbf{68.4} \\
\hline
\end{tabular}
\end{center}
\end{table}

\begin{table}[t]
\begin{center}
\caption{Per-class accuracy of Pascal VOC (\%) for 19+1}
\label{15-5}
\begin{tabular}{l|p{0.8cm}<{\centering}p{0.8cm}<{\centering}p{0.8cm}<{\centering}p{0.8cm}<{\centering}p{0.8cm}<{\centering}p{0.8cm}<{\centering}p{0.8cm}<{\centering}p{0.8cm}<{\centering}p{0.8cm}<{\centering}p{0.8cm}<{\centering}|c}
\hline
Methods & aero & bike & bird & boat & bottle & bus & car & cat & chair & cow & ~ \\
\hline
\hline
First 19 & 77.8 & 81.7 & 69.3 & 51.6 & 55.3 & 74.5 & 86.3 & 80.2 & 49.3 & 82.0 & ~ \\
New 1 & 0 & 0 & 0 & 0 & 0 & 0 & 0 & 0 & 0 & 0 & ~ \\
\hline
ILOD~\cite{shmelkov2017incremental} & 69.4 & 79.3 & 69.5 & 57.4 & 45.4 & 78.4 & 79.1 & 80.5 & 45.7 & 76.3 & ~ \\
Faster ILOD~\cite{peng2020faster} & 64.2 & 74.7 & 73.2 & 55.5 & 53.7 & 70.8 & 82.9 & 82.6 & 51.6 & 79.7 & ~ \\
\hline
ICOD(ours) & 71.6 & 73.7 & 69.8 & 58.2 & 41.7 & 71.5 & 73.2 & 72.0 & 57.6 & 75.1 & ~ \\
\hline
\hline
Methods & table & dog & horse & mbike & prsn & plant & sheep & sofa & train & tv & mAP \\\hline
\hline
First 19 & 63.6 & 76.8 & 80.9 & 77.5 & 82.4 & 42.9 & 73.9 & 70.4 & 70.4 & 0 & 67.3 \\
New 1 & 0 & 0 & 0 & 0 & 0 & 0 & 0 & 0 & 0 & 64.0 & 3.2 \\
\hline
ILOD~\cite{shmelkov2017incremental} & 64.8 & 77.2 & 80.8 & 77.5 & 70.1 & 42.3 & 67.5 & 64.4 & 76.7 & 62.7 & 68.3 \\
Faster ILOD~\cite{peng2020faster} & 58.7 & 78.8 & 81.8 & 75.3 & 77.4 & 43.1 & 73.8 & 61.7 & 69.8 & 61.1 & 68.6 \\
\hline
ICOD(ours) & 62.1 & 78.5 & 72.1 & 68.5 & 83.8 & 73.2 & 57.7 & 63.2 & 70.7 & 62.7 & 67.8 \\
\hline
\end{tabular}
\end{center}
\end{table}

\textbf{Pascal VOC}~\cite{everingham2010pascal} Pascal VOC is a famous benchmark for the object detection task, which annotates 20 different object categories. In total, Pascal VOC 2007 contains 9963 images collected from the Flickr photo-sharing website. Totally, 24640 objects are annotated in the dataset. In the experiment, 5K images for training and validation sets are used for training. The test split is used to get the evaluation results.

\textbf{Cityscapes}~\cite{cordts2016cityscapes} Cityscapes, which capture high-quality video in different cities, is collected for automotive vision. The dataset includes 5000 images which manually selected from 27 different cities. All images of Cityscapes are collected in good weather conditions. Because Cityscapes is designed for segmentation tasks, we generate the detection label following~\cite{chen2018domain}.

\textbf{Foggy Cityscapes}~\cite{sakaridis2018semantic} Foggy Cityscapes simulates the foggy weather based on the Cityscapes, which inherits the pixel-level labels of Cityscapes. Therefore, Cityscapes and Foggy Cityscapes share the same bounding boxes annotation. The validation set is also used for testing in the experiments.

\subsection{Incremental Learning of Different Categories}

For the setting of treating different categories as different tasks, following previous incremental object detection work~\cite{shmelkov2017incremental}, we implement our ICOD for the experiment.

\textbf{Experimental Setup} The experiment is based on the Pascal VOC dataset~\cite{everingham2010pascal}. Our model is implemented with the Res50 as the backbone for a fair comparison. The model is trained with the following three scenarios:
\begin{itemize}
\item We train the base model on 10 categories as the source task and learn new 10 categories for incremental learning (10+10).
\item The source task is to train with 15 categories, and the other 5 categories are used as the new task (15+5).
\item Firstly, we train 19 categories as the source task, and incrementally train the last category (19+1).
\end{itemize}
After training, we test our model with all 20 categories of the Pascal VOC dataset for comparison, where the problem of catastrophic forgetting can be observed.

\textbf{Implement Details} The causal features learned by our ICOD for the source task can be also used for other categories. In detail, we implement EWC~\cite{kirkpatrick2017overcoming} on the model to ensure the causal feature learned by the source task is not influenced by the new training process. The model is trained with 12 epochs, where the learning rate is changed from 0.001 to 0.0001 in the 8th epoch. Besides, we use some recent incremental object detection methods, including Faster-ILOD~\cite{peng2020faster} and ILOD~\cite{shmelkov2017incremental}, for comparison. The results of the experiment for the three different settings are shown in Table.~\ref{10-10}, Table.~\ref{15-5} and Table.~\ref{19-1}, respectively.

\textbf{Results} From the results, we can observe that our model achieves a great performance compared to other methods. Note that different from the other model which reviews the knowledge of the old task, our model freeze the backbone network and adapts to the new task based on the causal features. Thus, a better incremental learning speed can be achieved by our ICOD. Otherwise, the experiment also testifies the adaptability of the learned causal feature in our model. The data bias feature, which is learned in the traditional incremental object detection task, may not influence the performance of both the new task and the old task too much according to the results. The effectiveness of our model to focus on causal features and ignore data bias features is verified.

\subsection{Incremental Learning of Different Scenes}

Training the detector with different scenes or domains also makes the model lose some important information learned from old tasks. When we want to train a detector to adapt to a new scene or domain, the performance of the old task will drop. In this part, we will search the performance of our model with different kinds of scenes or domains. In detail, we implement Cityscapes as the source task, while the foggy cityscapes as the second task. Note that the two datasets annotate the same categories so that the experiment does not meet the condition of the new category.

\textbf{Implement Details} In this experiment, we firstly train our model on the source task (Cityscapes dataset). Then, like the experiments before, we fix the parameters of the backbone of the old model to make sure the causal features do not be affected, and finetune the classifier and regressor for the new task. VGG16 is implemented as the backbone of the experiment. Considering the sample number of category \emph{train} is too small, which may lead to unstable results, we exclude this category for our experiment. The results are presented in Table~\ref{f_city}, where \emph{city} stands for cityscapes, while \emph{fcity} is the foggy cityscapes.

\textbf{Results} According to the results, for the source task, our ICOD gets a great performance on both the cityscapes and foggy cityscapes datasets compared to the ordinary detector (i.e. Faster-RCNN). The adversarial learning strategy can help our model to enhance the effect of causal features and ignore the data bias features. With the robust casual features, ICOD can achieve better adaptability to other conditions or tasks, even for the unknown data. After finetuning the model with new data, better performance for both models can be achieved. With the causal feature, our model still gets better performance due to its robustness. The performance of the old task drops for all models, but ICOD achieves less influence. The experiment verifies that by learning the causal features, the robustness of the model is improved.

\begin{table}[t]
\begin{center}
\caption{Incremental learning from Cityscapes and Foggy Cityscapes.}
\label{f_city}
\begin{tabular}{p{2cm}|p{1cm}|p{0.9cm}<{\centering}p{0.9cm}<{\centering}p{0.9cm}<{\centering}p{0.9cm}<{\centering}p{0.9cm}<{\centering}p{0.9cm}<{\centering}p{0.9cm}<{\centering}||p{0.9cm}<{\centering}}
\hline
\multicolumn{2}{c|}{Old model} & person & rider & car & truck & bus & mcycle & bcycle & mAP \\
\hline
\hline
\multirow{2}*{FRCNN} & city & 41.7 & 55.0 & 58.2 & 35.2 & 58.4 & 39.3 & 44.2 & 47.4 \\
~ & fcity & 22.9 & 31.1 & 29.3 & 11.9 & 23.9 & 15.1 & 24.5 & 22.7 \\
\hline
\multirow{2}*{ICOD} & city & 42.4 & 54.6 & 59.1 & 37.6 & 60.7 & 42.4 & 46.0 & 49.0\\
~ & fcity & 26.4 & 37.3 & 35.5 & 15.6 & 26.5 & 23.1 & 32.2 & 28.1 \\
\hline
\hline
\multicolumn{2}{c|}{Finetuned} & person & rider & car & truck & bus & mcycle & bcycle & mAP \\
\hline
\multirow{2}*{FRCNN} & city & 44.6 & 53.3 & 53.8 & 32.9 & 54.5 & 37.2 & 41.8 & 44.6 \\
~ & fcity & 31.5 & 44.3 & 43.2 & 18.9 & 31.3 & 27.3 & 34.1 & 32.9 \\
\hline
\multirow{2}*{ICOD} & city & 41.4 & 53.7 & 58.9 & 39.6 & 60.8 & 41.0 & 42.9 & 48.3 \\
~ & fcity & 33.4 & 46.2 & 44.9 & 21.4 & 38.9 & 25.9 & 36.4 & 35.3 \\
\hline
\end{tabular}
\end{center}
\end{table}

\subsection{Model Analysis}

\begin{figure}[t]
\centering
\subfigure[Original Feature]
{\includegraphics[width=0.321\linewidth]{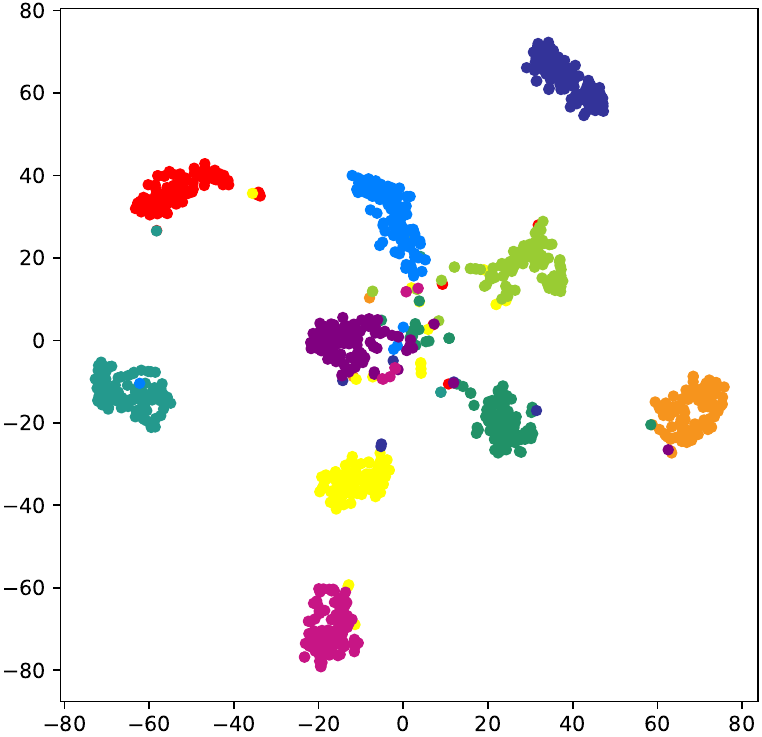}}
\subfigure[Causal Feature]
{\includegraphics[width=0.32\linewidth]{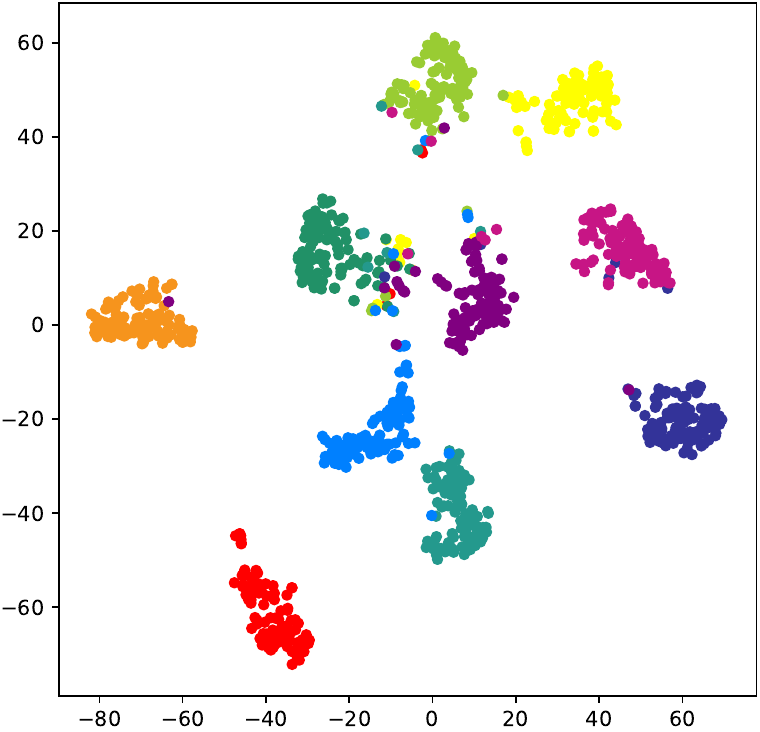}}
\subfigure[Bias Feature]
{\includegraphics[width=0.322\linewidth]{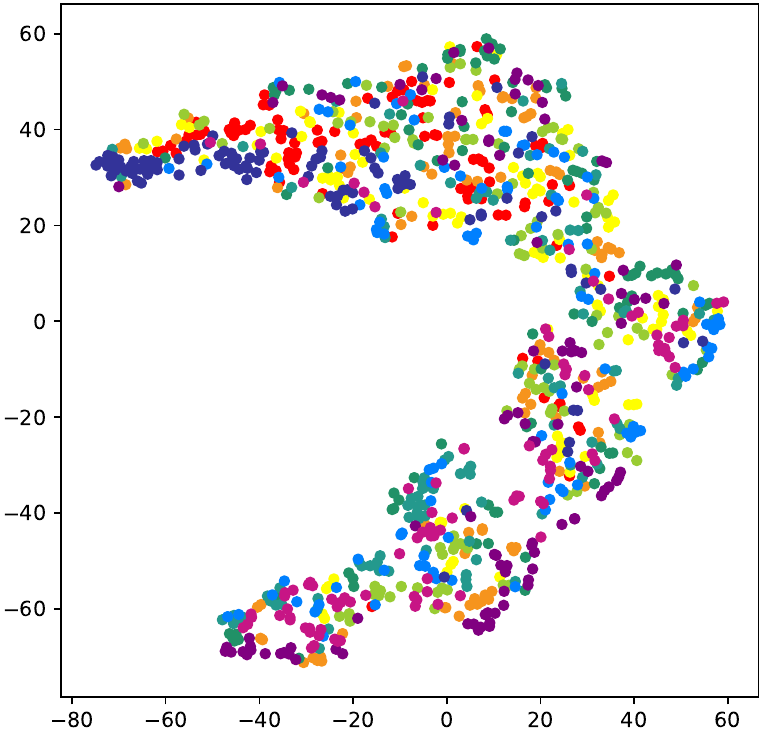}}
\caption{The t-SNE plot of the features in ICOD, where the original feature ($F$), causal feature ($F_{c}$), and bias feature ($F_{b}$) are presented. The visualization is based on the pascal voc 10+10 task. Features for the experiments are collected with the old model.}
\label{tsne}
\end{figure}

\textbf{Feature distribution of ICOD} In this part, we analyze the feature distribution of our model with t-SNE~\cite{van2008visualizing} plot. The original feature ($F$), causal feature ($F_{c}$), and data bias feature ($F_{b}$) are presented in the experiment. In detail, we crop the ground truth of the first 10 categories (source task), and then fed them into the source task model to get the features of instance level. Finally, we randomly select 100 samples for each category to get the t-SNE plot. The results are presented in Fig.~\ref{tsne}. Note that we ignore the random weight $r$ to get the causal features in the experiment.

From the results of t-SNE, we can get two points of conclusion: First, The causal features and original features get great feature distribution where features from different categories are clustered. On the contrary, the data bias feature is chaotic. Therefore, the effectiveness of the causal features can be observed. Second, the distribution of causal or original features is similar, which proves that our model ignores the data bias feature and focuses on the causal features to get the detection result. Besides, the unreliable relationship between the data bias feature and detection results is also suppressed.

\textbf{Why freeze the backbone.} In the task of cityscapes and foggy cityscapes, we freeze the backbone to research for the effectiveness of the causal features. Obviously, by learning the causal feature, the adaptability of the model improves. In this part, we search on the task by using EWC~\cite{kirkpatrick2017overcoming} as the substitute. The weight for EWC ranges from 0.01 to 0.5. Besides, we also implement the results of retraining (weight=0) for comparison. The results are presented in Table~\ref{f_ewc}.

\begin{table}[t]
\begin{center}
\caption{Incremental learning from Cityscapes and Foggy Cityscapes.}
\label{f_ewc}
\begin{tabular}{p{2cm}|p{1cm}|p{0.9cm}<{\centering}p{0.9cm}<{\centering}p{0.9cm}<{\centering}p{0.9cm}<{\centering}p{0.9cm}<{\centering}p{0.9cm}<{\centering}p{0.9cm}<{\centering}||p{0.9cm}<{\centering}}
\hline
weight & ~ & person & rider & car & truck & bus & mcycle & bcycle & mAP \\
\hline
\hline
\multirow{2}*{0.0} & city & 39.8 & 49.8 & 52.7 & 28.3 & 52.1 & 37.3 & 42.3 & 43.2
 \\
~ & fcity & 40.9 & 51.6 & 53.7 & 36.5 & 57.3 & 36.8 & 41.9 & 45.2\\
\hline
\multirow{2}*{0.01} & city & 40.8 & 52.0 & 53.6 & 33.3 & 54.6 & 35.7 & 41.5 & 44.5 \\
~ & fcity & 41.0 & 52.4 & 53.5 & 30.8 & 49.9 & 39.4 & 43.1 & 44.3 \\
\hline
\multirow{2}*{0.1} & city & 40.9 & 52.6 & 53.6 & 32.5 & 53.2 & 36.6 & 42.5 & 44.6 \\
~ & fcity & 41.2 & 51.5 & 53.5 & 30.9 & 51.4 & 42.0 & 44.1 & 45.0 \\
\hline
\multirow{2}*{0.5} & city & 40.7 & 50.7 & 53.7 & 36.3 & 51.9 & 33.6 & 39.8 & 44.7 \\
~ & fcity & 40.8 & 51.9 & 53.8 & 32.6 & 51.7 & 43.3 & 42.9 & 45.3\\
\hline
\end{tabular}
\end{center}
\end{table}

Obviously, by training more parameters compare to the experiment of Table.~\ref{f_city}, the model can get better performance on foggy cityscapes. With EWC as a substitute, the model can learn some features of other tasks during the training phase. Otherwise, the experiment shows an interesting result. First, EWC does not benefit the catastrophic forgetting problem too much. Second, as the weight of EWC increases, the parameters of the old model are harder to be changed, but the performance of foggy cityscapes is also increased. We think it might be because some useful information learned in the source task is also valid for the new task, which even can not be learned with the new task data. That is, preserving the features of the source task can also benefit the new task. In our ICOD, our model can learn the causal features, by preserving these features for training the new task, the performance can also be improved while prevent the catastrophic forgetting problem.

\subsection{Visualization of Detection Results.}

\begin{figure}[t]
\centering
\includegraphics[width=1.0\linewidth]{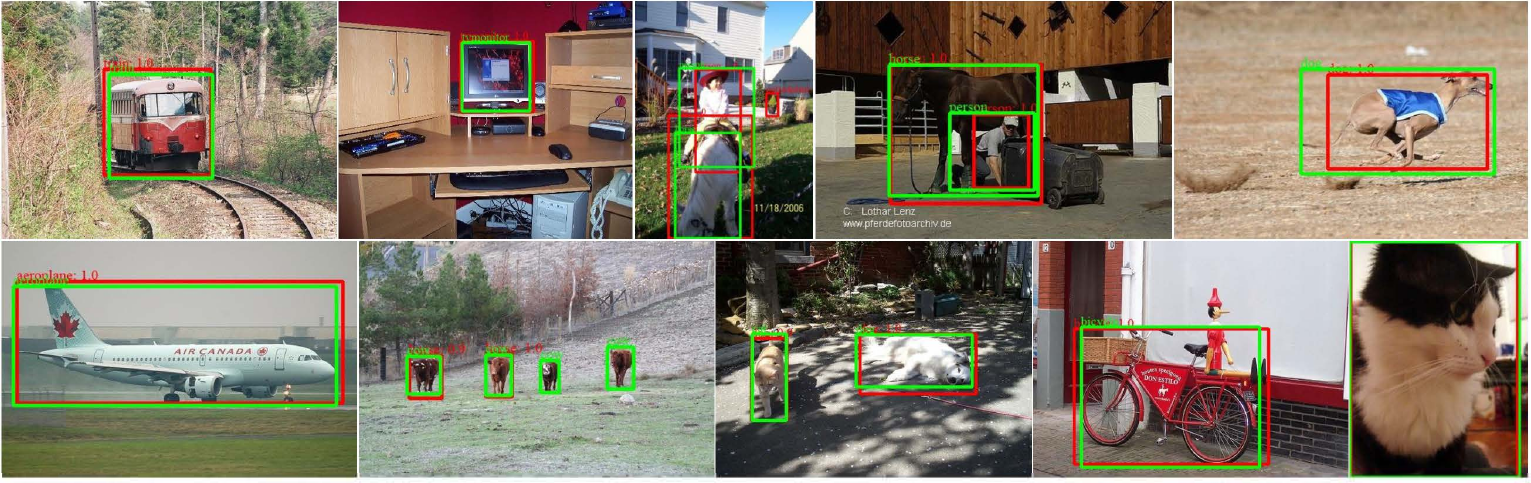}
\caption{The visualization of Detection Results. The ground truth is presented in green boxes, while the detection results are shown in red. Objects from both the new and old categories are presented.}
\label{vis}
\end{figure}

In this part, we will show some visualization results of our ICOD, which are generated with a new task detection model based on Pascal VOC (10-10task). The detected results are shown in the red box, while the ground truth is presented in the green box. The results are shown in Fig.~\ref{vis}. According to the results, we can observe that the model can detect the categories from both the old and new tasks. Besides, our model still meets some flaws according to the visualization, in some samples, the categories of the old task are likely to be recognized as a similar category of the new task. Some of the objects of the new task categories are missed for detection. This phenomenon is the reflection of the catastrophic forgetting problem, which might be our future work.

\section{Conclusion}

Conventional object detection model is restricted by its training set. Users often require them to extend the application scenario of the detection model in practice. However, the detection model often meets the catastrophic forgetting problem when learning the new task. The traditional incremental object detection model focuses on reviewing the knowledge of the old task with knowledge distillation, which ignores feature learning. In this paper, we introduce the incremental causal object detection (ICOD) model for the incremental object detection task. Different from traditional models, our ICOD aims to learn more robust features (i.e. causal feature) for different tasks. In detail, by enhancing the effectiveness of the causal features and suppressing the data bias feature, our model gets better adaptability to other tasks. We explain the causal preliminaries as the theoretical basis of our model and conduct several experiments to show the effectiveness of our ICOD.

\clearpage
\bibliographystyle{splncs04}
\bibliography{egbib}
\end{document}